\documentclass[authoryear,final,5p,times,twocolumn]{elsarticle}

\usepackage{graphics}

\usepackage{amssymb}

\usepackage{multirow}

\journal{Expert Systems with Applications}

\renewcommand{\figurename}{Fig.}
\newcommand{\figref}[1]{\figurename~\ref{#1}}

\begin{document}

\begin{frontmatter}



\title{An expert system for detecting automobile insurance fraud using social network analysis}


\cortext[cor]{Corresponding author. Tel.: +386 1 4768 186.}
\author{Lovro \v Subelj\corref{cor}}
\ead{lovro.subelj@fri.uni-lj.si}
\author{\v Stefan Furlan}
\ead{stefan.furlan@fri.uni-lj.si}
\author{Marko Bajec}
\ead{marko.bajec@fri.uni-lj.si}
\address{Faculty of Computer and Information Science, University of Ljubljana, Tr\v za\v ska 25, SI-1001 Ljubljana, Slovenia}

\begin{abstract}
The article proposes an expert system for detection, and subsequent investigation, of groups of collaborating automobile insurance fraudsters. The system is described and examined in great detail, several technical difficulties in detecting fraud are also considered, for it to be applicable in practice. 
Opposed to many other approaches, the system uses networks for representation of data. Networks are the most natural representation of such a relational domain, allowing formulation and analysis of complex relations between entities. Fraudulent entities are found by employing a novel assessment algorithm, \textit{Iterative Assessment Algorithm} (\textit{IAA}), also presented in the article. Besides intrinsic attributes of entities, the algorithm explores also the relations between entities.
The prototype was evaluated and rigorously analyzed on real world data. 
Results show that automobile insurance fraud can be efficiently detected with the proposed system and that appropriate data representation is vital. 
\end{abstract}


\end{frontmatter}


\section{Introduction}
\label{introduction}
Fraud is encountered in a variety of domains. It comes in all different shapes and sizes, from traditional fraud, e.g. (simple) tax cheating, to more sophisticated, where entire \textit{groups} of individuals are collaborating in order to commit fraud. Such groups can be found in the automobile insurance domain.

Here fraudsters stage traffic accidents and issue fake insurance claims to gain (unjustified) funds from their general or vehicle insurance. There are also cases where an accident has never occurred, and the vehicles have only been placed onto the road. Still, the majority of such fraud is not planned (\textit{opportunistic fraud}) \textendash\space an individual only seizes the opportunity arising from the accident and issues exaggerated insurance claims or claims for past damages.

Staged accidents have several common characteristics. They occur in late hours and non-urban areas in order to reduce the probability of witnesses. Drivers are usually younger males, there are many passengers in the vehicles, but never children or elders. The police is always called to the scene to make the subsequent acquisition of means easier. It is also not uncommon that all of the participants have multiple (serious) injuries, when there is almost no damage on the vehicles. Many other suspicious characteristics exist, not mentioned here.

The insurance companies place the most interest in organized groups of fraudsters consisting of drivers, chiropractors, garage mechanics, lawyers, police officers, insurance workers and others. Such groups represent the majority of revenue leakage. Most of the analyses agree that approximately $20\%$ of all insurance claims are in some way fraudulent (various resources). But most of these claims go unnoticed, as fraud investigation is usually done by hand by the domain expert or investigator and is only rarely computer supported. Inappropriate representation of data is also common, making the detection of groups of fraudsters extremely difficult. An expert system approach is thus needed.

\citet{Jen97} has observed several technical difficulties in detecting fraud (various domains). Most hold for (automobile) insurance fraud as well. Firstly, only a small portion of accidents or participants is fraudulent (\textit{skewed class distribution}) making them extremely difficult to detect. Next, there is a severe lack of \textit{labeled} data sets as labeling is expensive and time consuming. Besides, due to sensitivity of the domain, there is even a lack of unlabeled data sets. Any approach for detecting such fraud should thus be founded on moderate resources (data sets) in order to be applicable in practice. Fraudsters are very innovative and new types of fraud emerge constantly. Hence, the approach must also be highly adaptable, detecting new types of fraud as soon as they are noticed. Lastly, it holds that fully autonomous detection of automobile insurance fraud is not possible in practice. Final assessment of potential fraud can only be made by the domain expert or investigator, who also determines further actions in resolving it. The approach should also support this investigation process.

Due to everything mentioned above, the set of approaches for detecting such fraud is extremely limited. We propose a novel expert system approach for detection and subsequent investigation of automobile insurance fraud. The system is focused on detection of groups of collaborating fraudsters, and their connecting accidents (non-opportunistic fraud), and not some isolated fraudulent entities. The latter should be done independently for each particular entity, while in our system, the entities are assessed in a way that considers also the relations between them. This is done with appropriate representation of the domain \textendash\space networks.

Networks are the most natural representation of any relational domain, allowing formulation of complex relations between entities. They also present the main advantage of our system against other approaches that use a standard \textit{flat data} form. As collaborating fraudsters are usually related to each other in various ways, detection of groups of fraudsters is only possible with appropriate representation of data. Networks also provide clear visualization of the assessment, crucial for the subsequent investigation process.

The system assesses the entities using a novel \textit{Iterative Assessment Algorithm} (\textit{IAA} algorithm), presented in this article. No learning from initial labeled data set is done, the system rather allows simple incorporation of the domain knowledge. This makes it applicable in practice and allows detection of new types of fraud as soon as they are encountered. The system can be used with poor data sets, which is often the case in practice. To simulate realistic conditions, the discussion in the article and evaluation with the prototype system relies only on the data and entities found in the police record of the accident (main entities are participant, vehicle, collision\footnote{Throughout the article the term collision is used instead of (traffic) accident. The word accident implies there is no one to blame, which contradicts with the article.}, police officer).

The article makes an in depth description, evaluation and analysis of the proposed system. We pursue the hypothesis that automobile insurance fraud can be detected with such a system and that proper data representation is vital. Main contributions of our work are: (1) a novel expert system approach for the detection of automobile insurance fraud with networks; (2) a benchmarking study, as no expert system approach for detection of groups of automobile insurance fraudsters has yet been reported (to our knowledge); (3) an algorithm for assessment of entities in a relational domain, demanding no labeled data set (\textit{IAA} algorithm); and (4) a framework for detection of groups of fraudsters with networks (applicable in other relational domains).

The rest of the article is organized as follows. In section~\ref{related_work} we discuss related work and emphasize weaknesses of other proposed approaches. Section~\ref{background} presents formal grounds of (social) networks. Next, in section~\ref{system}, we introduce the proposed expert system for detecting automobile insurance fraud. The prototype system was evaluated and rigorously analyzed on real world data, description of the data set and obtained results are given in section~\ref{evaluation}. Discussion of the results is conducted in section~\ref{discussion}, followed by the conclusion in section~\ref{conclusion}.

\section{Related work}
\label{related_work}
Our work places in the wide field of fraud detection. Fraud appears in many domains including telecommunications, banking, medicine, e-commerce, general and automobile insurance. Thus a number of expert system approaches for preventing, detecting and investigating fraud have been developed in the past. Researches have proposed using some standard methods of data mining and machine learning, \textit{neural networks}, \textit{fuzzy logic}, \textit{genetic algorithms}, \textit{support vector machines}, \textit{(logistic) regression}, \textit{consolidated (classification) trees}, approaches over \textit{red-flags} or \textit{profiles}, various statistical methods and other methods and approaches \citep{AAG02,BDGLA02,BH02,EHP06,FB08,GS99,HLV07,KSM07,PMAGM05,RKK07,QS08,SVCS09,VDBD02,VDD05,WD98,YH06}. Analyses show that in practice none is significantly better than others \citep{BH02,VDD05}. Furthermore, they mainly have three weaknesses. They (1) use inappropriate or inexpressive representation of data; (2) demand a labeled (initial) data set; and (3) are only suitable for larger, richer data sets. It turns out that these are generally a problem when dealing with fraud detection \citep{Jen97,PLSG05}.

In the narrower sense, our work comes near the approaches from the field of network analysis, that combine intrinsic attributes of entities with their relational attributes. \citet{NC03} proposed detecting anomalies in networks with various types of vertices, but they focus on detecting suspicious structures in the network, not vertices (i.e. entities). Besides that, the approach is more appropriate for larger networks. Researchers also proposed detecting anomalies using measures of centrality \citep{Fre77,Fre79}, random walks \citep{SQCF05} and other \citep{HC03,MT00}, but these approaches mainly rely only on the relational attributes of entities.

Many researchers have investigated the problem of classification in the relational context, following the hypothesis that classification of an entity can be improved by also considering its related entities (inference). Thus many approaches formulating \textit{inference}, \textit{spread} or \textit{propagation} on networks have been developed in various fields of research \citep{BP98,DR01,Kle99,KF98,LG03a,Min01,NJ00}. Most of them are based on one of the three most popular (approximate) inference algorithms: \textit{Relaxation Labeling (RL)}~\citep{HZ83} from the computer vision community, \textit{Loopy Belief Propagation (LBP)} on loopy (Bayesian) \textit{graphical models} \citep{KF98} and \textit{Iterative Classification Algorithm (ICA)} from the data mining community \citep{NJ00}. For the analyses and comparison see~\citep{KKT03,SG07}.

Researchers have reported good results with these algorithms \citep{BP98,KF98,LG03a,NJ00}, however they mainly address the problem of learning from an (initial) labeled data set (\textit{supervised learning}), or a partially labeled (\textit{semi-supervised learning}) \citep{LG03b}, therefore the approaches are generally inappropriate for fraud detection. The algorithm we introduce here, \textit{IAA} algorithm, is almost identical to the \textit{ICA} algorithm, however it was developed with different intentions in mind \textendash\space to assess the entities when no labeled data set is at hand (and not for improving classification with inference). Furthermore, \textit{IAA} does not address the problem of \textit{classification}, but \textit{ranking}. Thus, in this way, it is actually a simplification of \textit{RL} algorithm, or even Google's \textit{PageRank}~\citep{BP98}, still it is not founded on the probability theory like the latter.

We conclude that due to the weaknesses mentioned, most of the proposed approaches are inappropriate for detection of (automobile) insurance fraud. Our approach differs, as it does not demand a labeled data set and is also appropriate for smaller data sets. It represents data with networks, which are one of the most natural representation and allow complex analysis without simplification of data. It should be pointed out that networks, despite their strong foundations and expressive power, have not yet been used for detecting (automobile) insurance fraud (at least according to our knowledge).

\section{(Social) networks}
\label{background}
\begin{figure}[htp]
\begin{center}
\includegraphics[width=1.\columnwidth]{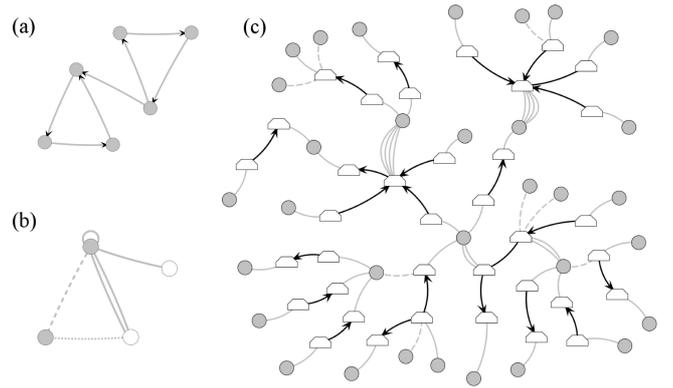} 
\caption{(a) simple graph with directed edges; (b) undirected multigraph with labeled vertices and edges (labels are represented graphically); (c) network representing collisions where round vertices correspond to participants and cornered vertices correspond to vehicles. Collisions are represented with directed edges between vehicles.}
\label{fig:graphs}
\end{center}
\end{figure}

Networks are based upon mathematical objects called \textit{graphs}. Informally speaking, graph consists of a collection of points, called \textit{vertices}, and links between these points, called \textit{edges} (\figref{fig:graphs}). Let $V_G$, $E_G$ be a set of vertices, edges for some graph $G$ respectively. We define $G$ as $G=(V_G,E_G)$ where
\begin{eqnarray}
V_G & = & \{v_1,v_2\dots v_n\}, \\
E_G & \subseteq & \{\{v_i,v_j\}|\mbox{ }v_i,v_j\in V_G\wedge i\neq j\}.
\label{eq_E_undirected}
\end{eqnarray}
Note that edges are sets of vertices, hence they are not directed (\textit{undirected graph}). In the case of \textit{directed graphs} equation~(\ref{eq_E_undirected}) rewrites to
\begin{eqnarray}
E_G & \subseteq & \{(v_i,v_j)|\mbox{ }v_i,v_j\in V_G\wedge i\neq j\},
\label{eq_E_directed}
\end{eqnarray}
where edges are ordered pairs of vertices \textendash\space $(v_i,v_j)$ is an edge from $v_i$ to $v_j$. The definition can be further generalized by allowing multiple edges between two vertices and loops (edges that connect vertices with themselves). Such graphs are called \textit{multigraphs}. Examples of some simple (multi)graphs can be seen in \figref{fig:graphs}.

In practical applications we usually strive to store some extra information along with the vertices and edges. Formally, we can define two labeling functions 
\begin{eqnarray}
l_{V_G}: & V_G\rightarrow\Sigma_{V_G}, \\
l_{E_G}: & E_G\rightarrow\Sigma_{E_G},
\end{eqnarray}
where $\Sigma_{V_G}$, $\Sigma_{E_G}$ are (finite) alphabets of all possible vertex, edge labels respectively. \textit{Labeled graph} can be seen in \figref{fig:graphs}~(b).

We proceed by introducing some terms used later on. Let $G$ be some undirected multigraph or an \textit{underlying graph} of some directed multigraph \textendash\space underlying graph consists of same vertices and edges as the original directed (multi)graph, only that all of its edges are set to be undirected. $G$ naturally partitions into a set of \textit{(connected) components} denoted $C(G)$. E.g. all three graphs in \figref{fig:graphs} have one connected component, when graphs in \figref{fig:system} consist of several connected components. From here on, we assume that $G$ consists of a single connected component.

Let $v_i$ be some vertex in graph $G$, $v_i\in V_G$. \textit{Degree} of the vertex $v_i$, denoted $d(v_i)$, is the number of edges incident to it. Formally,
\begin{eqnarray}
d(v_i) & = & |\{e|\mbox{ }e\in E_G\wedge v_i\in e\}|.
\end{eqnarray}
Let $v_j$ be some other vertex in graph $G$, $v_j\in V_G$, and let $p(v_i,v_j)$ be a \textit{path} between $v_i$ and $v_j$. A path is a sequence of vertices on a way that leads from one vertex to another (including $v_i$ and $v_j$). There can be many paths between two vertices. A \textit{geodesic} $g(v_i,v_j)$ is a path that has the minimum size \textendash\space consists of the least number of vertices. Again, there can also be many geodesics between two vertices. 

We can now define the \textit{distance} between two vertices, i.e. $v_i$ and $v_j$, as
\begin{eqnarray}
d(v_i,v_j) & = & |g(v_i,v_j)|-1.
\end{eqnarray}
Distance between $v_i$ and $v_j$ is the number of edges visited when going from $v_i$ to $v_j$ (or vice versa). The \textit{diameter} of some graph $G$, denoted $d(G)$, is a measure for the ``width'' of the graph. Formally, it is defined as the maximum distance between any two vertices in the graph,
\begin{eqnarray}
d(G) & = & \max\{d(v_i,v_j)|\mbox{ }v_i,v_j\in V_G\}.
\end{eqnarray}

All graphs can be divided into two classes. First are \textit{cyclic} graphs, having a path $p(v_i,v_i)$ that contains at least two other vertices (besides $v_i$) and has no repeated vertices. Such path is called a \textit{cycle}. Graphs in \figref{fig:graphs}~(a)~and~(b) are both cyclic. Second class of graphs consists of \textit{acyclic} graphs, more commonly known as \textit{trees}. These are graphs that contain no cycle (see \figref{fig:graphs}~(c)). Note that a simple undirected graph is a tree if and only if $|E_G|=|V_G|-1$.

Finally, we introduce the \textit{vertex cover} of a graph $G$. Let $S$ be a subset of vertices, $S\subseteq V_G$, with a property that each edge in $E_G$ has at least one of its incident vertices in $S$ (covered by $S$). Such $S$ is called a vertex cover. It can be shown, that finding a minimum vertex cover is \textit{NP-hard} in general.

Graphs have been studied and investigated for almost $300$ years thus a strong theory has been developed until today. There are also numerous practical problems and applications where graphs have shown their usefulness \citep[e.g.][]{BP98} \textendash\space they are the most natural representation of many domains and are indispensable whenever we are interested in relations between entities or in patterns in these relations. We emphasize this only to show that networks have strong mathematical, and also practical, foundation \textendash\space \textit{networks}\footnote{Throughout the article the terms graph and network are used as synonyms.} are usually seen as labeled, or \textit{weighted}, multigraphs with both directed and undirected edges (see \figref{fig:graphs}~(c)). Furthermore, vertices of a network usually represent some entities, and edges represent some relations between them. When vertices correspond to people, or groups of people, such networks are called \textit{social networks}.

Networks often consist of densely connected subsets of vertices called \textit{communities}. Formally, communities are subsets of vertices with many edges between the vertices within some community and only a few edges between the vertices of different communities. \citet{GN02} suggested identifying communities by recursively removing the edges between them \textendash\space \textit{between edges}. As it holds that many geodesics run along such edges, where only few geodesics run along edges within communities, between edges can be removed by using \textit{edge betweenness} \citep{GN02}. It is defined as
\begin{eqnarray}
Bet(e_i) & = & |\{g(v_i,v_j)|\mbox{ }v_i,v_j\in V_G\wedge \\
& & \wedge\mbox{ }g(v_i,v_j)\mbox{ goes along }e_i\}|,
\nonumber
\end{eqnarray}
where $e_i\in E_G$. The edge betweenness $Bet(e_i)$ is thus the number of all geodesics that run along edge $e_i$. 

For more details on (social) networks see e.g. \citep{New03,New08}.

\section{Expert system for detecting automobile insurance fraud}
\label{system}
\begin{figure}[htp]
\begin{center}
\includegraphics[width=0.30\columnwidth]{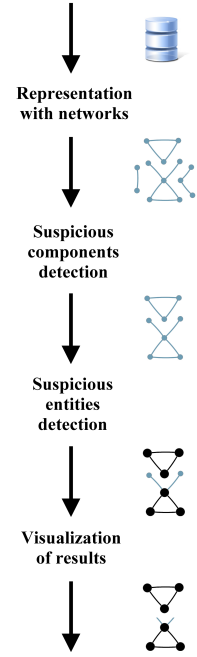} 
\caption{Framework of the proposed expert system for detecting (automobile insurance) fraud.}
\label{fig:system}
\end{center}
\end{figure}

As mentioned above, the proposed expert system uses (primarily constructed) networks of collisions to assign suspicion score to each entity. These scores are used for the detection of groups of fraudsters and their corresponding collisions. The \textit{framework} of the system is structured into four \textit{modules} (\figref{fig:system}).

In the first module, different types of networks are constructed from the given data set. When necessary, the networks are also simplified \textendash\space divided into natural communities that appear inside them. The latter is done without any loss of generality. 

Networks from the first module naturally partition into several connected components. In the second module we investigate these components and output the suspicious, focusing mainly on their structural properties such as diameter, cycles, etc. Other components are discarded at the end of this module.

Not all entities in some suspicious component are necessarily suspicious. In the third module components are thus further analyzed in order to detect key entities inside them. They are found by employing \textit{Iterative Assessment Algorithm (IAA)}, presented in this article. The algorithm assigns a suspicion score to each entity, which can be used for subsequent assessment and analysis \textendash\space to identify suspicious groups of entities and their connecting collisions. In general, suspicious groups are subsets of suspicious components.

Note that detection of suspicious entities is done in two \textit{stages} (second and third module). In the first stage, or the second module, we focus only on detecting suspicious components and in the second stage, third module, we also locate the suspicious entities within them. Hence the detection in the first, second stage is done at the level of components, entities respectively. The reason for this \textit{hierarchical investigation} is that early stages simplify assessment in the later stages, possibly without any loss for detection (for further implications see section~\ref{discussion}).

%
%

It holds that fully autonomous detection of automobile insurance fraud is not possible in practice. The obtained results should always be investigated by the domain expert or investigator, who determines further actions for resolving potential fraud. The purpose of the last, fourth, module of the system is thus to appropriately assess and visualize the obtained results, allowing the domain expert or investigator to conduct subsequent analysis.

First three modules of the system are presented in sections~\ref{system_representation}, \ref{system_components},~\ref{system_entities} respectively, when the last module is only briefly discussed in section~\ref{system_remarks}.

\subsection{Representation with networks}
\label{system_representation}
Every entity's attribute is either \textit{intrinsic} or \textit{relational}. Intrinsic attributes are those, that are independent of the entity's surrounding (e.g. person's age), while the relational attributes represent, or are dependent on, relations between entities (e.g. relation between two colliding drivers). Relational attributes can be naturally represented with the edges of a network. Thus we get networks, where vertices correspond to entities and edges correspond to relations between them. Numerous different networks can be constructed, depending on which entities we use and how we connect them to each other.

The purpose of this first module of the system is to construct different types of networks, used later on. It is not immediately clear how to construct networks, that describe the domain in the best possible way and are most appropriate for our intentions. This problem arises as networks, despite their high expressive power, are destined to represent relations between only two entities (i.e. \textit{binary relations}). As collisions are actually relations between multiple entities, some sort of projection of the data set must be made (for other suggestions see section~\ref{conclusion}).

Collisions can thus be represented with various types of networks, not all equally suitable for fraud detection. In our opinion, there are some guidelines that should be considered when constructing networks from any relational domain data (guidelines are given approximately in the order of their importance):
\begin{enumerate}[1.]
\item \textit{Intention:} networks should be constructed so that they are most appropriate for our intentions (e.g. fraud detection)
\label{guideline_intentions}
\item \textit{Domain:} networks should be constructed in a way that describes the domain as it is (e.g. connected vertices should represent some entities, also directly connected in the data set)
\label{guideline_domain}
\item \textit{Expressiveness:} expressive power of the constructed networks should be as high as possible
\label{guideline_expressiveness}
\item \textit{Structure:} structure of the networks should not be used for describing some specific domain characteristics (e.g. there should be no cycles in the networks when there are no actual cycles in the data set). Structural properties of networks are a strong tool that can be used in the subsequent (investigation) process, but only when these properties were not artificially incorporated into the network during the construction process
\label{guideline_structure}
\item \textit{Simplicity:} networks should be kept as simple and sparse as possible (e.g. not all entities need to be represented by its own vertices). The hypothesis here is that simple networks would also allow simpler subsequent analysis and clearer final visualization (principle of \textit{Occam's razor}\footnote{The principle states that the explanation of any phenomenon should make as few assumptions as possible, eliminating those making no difference in the assessment \textendash\space entities should not be multiplied beyond necessity.})
\label{guideline_simplicity}
\item \textit{Uniqueness:} every network should uniquely describe the data set being represented (i.e. there should be a \textit{bijection} between different data sets and corresponding networks)
\label{guideline_uniqueness}
\end{enumerate}
Frequently all guidelines can not be met and some trade-off have to be made. 

\begin{figure}[htp]
\begin{center}
\includegraphics[width=1.0\columnwidth]{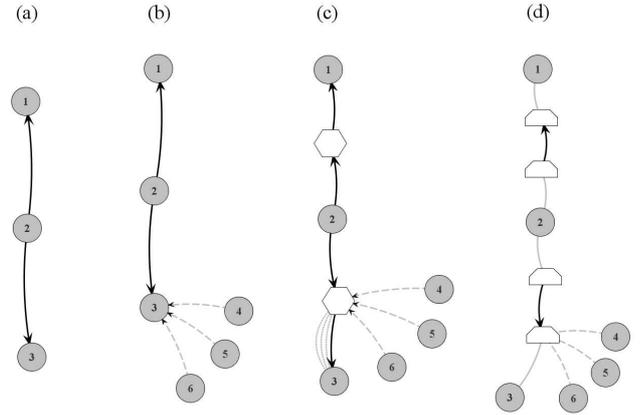} 
\caption{Four types of networks representing same two collisions \textendash\space (a) \textit{drivers network}, (b) \textit{participants network}, (c) \textit{COPTA network} and (d) \textit{vehicles network}. Rounded vertices correspond to participants, hexagons correspond to collisions and irregular cornered vertices correspond to vehicles. Solid directed edges represent involvement in some collision, solid undirected edges represent drivers (only for the vehicles network) and dashed edges represent passengers. Guilt in the collision is formulated with edge's direction.}
\label{fig:collisions_networks}
\end{center}
\end{figure}

In general there are ${3 \choose 1}+{3 \choose 2}+({3 \choose 2}+{3 \choose 3})=10$ possible ways how to connect three entities (i.e. collision, participant and vehicle), depending on which entities we represent with their own vertices. $7$ of these represent participants with vertices and in $4$ cases all entities are represented by their own vertices. For the reason of simplicity, we focus on the remaining $3$ cases. In the following we introduce four different types of such networks, as an example and for later use. All can be seen in \figref{fig:collisions_networks}.

The simplest way is to only connect the drivers who were involved in the same collision \textendash\space \textit{drivers networks}. Guilt in the collision is formulated with edge's direction. Note that drivers networks severely lack expressive power (guideline~\ref{guideline_expressiveness}). We can therefore add the passengers and get \textit{participants networks}, where passengers are connected with the corresponding drivers. Such networks are already much richer, but they have one major weakness  \textendash\space passengers ``group'' on the driver, i.e. it is generally not clear which passengers were involved in the same collision and not even how many passengers were involved in some particular collision (guidelines~\ref{guideline_expressiveness},~\ref{guideline_uniqueness}).

This weakness is partially eliminated by \textit{COnnect Passengers Through Accidents networks} (\textit{COPTA networks}). We add special vertices representing collisions and all participants in some collision are now connected through these vertices. Passengers no longer group on the drivers but on the collisions, thus the problem is partially eliminated. We also add special edges between the drivers and the collisions, to indicate the number of passengers in the vehicle. This type of networks could be adequate for many practical applications, but it should be mentioned that the distance between two colliding drivers is now twice as large as before \textendash\space the drivers are those that were directly related in the collision (guideline~\ref{guideline_domain},~\ref{guideline_simplicity}).

Last type of networks are \textit{vehicles networks} where special vertices are added to represent vehicles. Collisions are now represented by edges between vehicles, and driver and passengers are connected through them. Such networks provide good visualization of the collisions and also incorporate another entity, but they have many weaknesses as well. Two colliding drivers are very far apart and (included) vehicles are not actually of our interest (guideline~\ref{guideline_simplicity}). Such networks also seem to suggest that the vehicles are the ones, responsible for the collision (guideline~\ref{guideline_domain}). Vehicles networks are also much larger than the previous.

A better way to incorporate vehicles into networks is simply to connect collisions, in which the same vehicle was involved. Similar holds for other entities like police officers, chiropractors, lawyers, etc. Using special vertices for these entities would only unnecessarily enlarge the networks and consequently make subsequent detection harder (guidelines~\ref{guideline_intentions},~\ref{guideline_simplicity}). It is also true, that these entities usually aren't available in practice (sensitivity of the domain).

Summary of the analysis of different types of networks is given in table~\ref{tbl:networks_guidelines}.

\begin{table}[htp]
\begin{center}
\begin{tabular}{cccccc} 
\multicolumn{6}{l}{\textit{Guidelines and networks}} \\\hline\hline
 & \textit{drivers} & \textit{particip.} & \textit{COPTA} & \textit{vehicles} & \\\hline
\multirow{2}{*}{\textit{Intention}} & $+$ & $++$ & & & \multirow{2}{*}{$5$} \\
 & & $+$ & $++$ & & \\
\textit{Domain} & & & $-$ & $-$ & $4$ \\
\textit{Expressive.} & $--$ & $-$ & & $+$ & $4$ \\
\textit{Structure} & & & & & $4$ \\
\textit{Simplicity} & $+$ & & $-$ & $--$ & $3$ \\
\textit{Uniqueness} & & $-$ & $-$ & $-$ & $2$ \\\hline
\multirow{2}{*}{Total} & $0$ & $\mathbf{4}$ & $-9$ & $-8$ & \\
 & $-5$ & $-1$ & $\mathbf{1}$ & $-8$ & \\
\end{tabular}
\end{center}
\caption{Comparison of different types of networks due to the proposed guidelines. Scores assigned to the guidelines are a choice made by the authors. Analysis for \textit{Intention} (guideline~\ref{guideline_intentions}), and total score, is given separately for second, third module respectively.}
\label{tbl:networks_guidelines}
\end{table}

There is of course no need to use the same type of networks in every stage of the detection process (guideline~\ref{guideline_intentions}). In the prototype system we thus use participants networks in the second module (section~\ref{system_components}), as they provide enough information for initial suspicious components detection, and \textit{COPTA} networks in the third module (section~\ref{system_entities}), whose adequacy will be clearer later. Other types of networks are used only for visualization purposes. Network scores, given in table~\ref{tbl:networks_guidelines}, confirm this choice.

After the construction of networks is done, the resulting connected components can be quite large (depending on the type of networks used). As it is expected that groups of fraudsters are relatively small, the components should in this case be simplified. We suggest using edge betweenness~\citep{GN02} to detect communities in the network (i.e. supersets of groups of fraudsters) by recursively removing the edges until the resulting components are small enough. As using edge betweenness assures that we would be removing only the edges between the communities, and not the edges within communities, simplification is done without any loss for generality.

\subsection{Suspicious components detection}
\label{system_components}
The networks from the first module consist of several connected components. Each component describes a group of related entities (i.e. participants, due to the type of networks used), where some of these groups contain fraudulent entities. Within this module of the system we want to detect such groups (i.e. \textit{fraudulent components}) and discard all others, in order to simplify the subsequent detection process in the third module. Not all entities in some fraudulent component are necessarily fraudulent. The purpose of the third module is to identify only those that are. 

\begin{figure}[htp]
\begin{center}
\includegraphics[width=0.75\columnwidth]{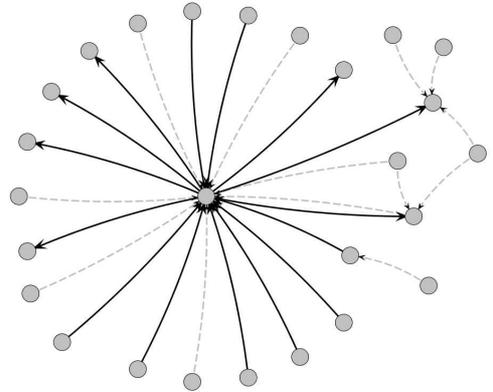} 
\caption{Example of a component of participants network with many of the suspicious characteristics shared by fraudulent components.}
\label{fig:suspicious_component}
\end{center}
\end{figure}

Analyses, conducted with the help of a domain expert, showed that fraudulent components share several \textit{structural characteristics}. Such components are usually much larger than other, non-fraudulent components, and are also denser. The underlying collisions often happened in suspicious circumstances, and the ratio between the number of collisions and the number of different drivers is usually close to $1$ (for reference, the ratio for completely independent collisions is $2$). There are vertices with extremely high degree and \textit{centrality}. Components have a small diameter, (short) cycles appear and the size of the minimum vertex cover is also very small (all due to the size of the component). There are also other characteristics, all implying that entities, represented by such components, are unusually closely related to each other. Example of a fraudulent component with many of the mentioned characteristics is shown in \figref{fig:suspicious_component}.

We have thus identified several \textit{indicators} of likelihood that some component is fraudulent (i.e. \textit{suspicious component}). The detection of suspicious components is done by assessing these indicators. Only simple indicators are used (no combinations of indicators).

Formally, we define an ensemble of $n$ indicators as $I = [I_1, I_1 \dots I_n]^T$. Let $c$ be some connected component in network $G$, $c \in C(G)$, and let $H_i(c)$ be the value for $c$ of the characteristic, measured by indicator $I_i$. Then
\begin{eqnarray}
I_i(c) & = & \left\{\begin{array}{cl}
1 & c \mbox{ has suspicious value of }H_i \\
0 & \mbox{otherwise}
\end{array}\right..
\label{eq_I_i}
\end{eqnarray}
For the reason of simplicity, all indicators are defined as \textit{binary attributes}. For the indicators that measure a characteristic that is independent of the structure of the component (e.g. number of vertices, collisions, etc.), simple \textit{thresholds} are defined in order to distinguish suspicious components from others (due to this characteristic). These thresholds are set by the domain expert.

Other characteristics are usually greatly dependent on the number of the vertices and edges in the component. A simple \textit{threshold strategy} thus does not work. Values of such $H_i$ could of course be ``normalized'' before the assessment (based on the number of vertices and edges), but it is often not clear how. Values could also be assessed using some (supervised) learning algorithm over a labeled data set, but a huge set would be needed, as the assessment should be done for each number of vertices and edges separately (owing to the dependence mentioned). What remains is to construct random networks of (presumably) honest behavior and assess the values of such characteristics using them.

No in-depth analysis of collisions networks has so far been reported, and it is thus not clear how to construct such random networks. General random network \textit{generators} or \textit{models}, e.g. \citep{BA99,EW02}, mainly give results far away from the collisions networks (visually and by assessing different characteristics). Therefore a sort of \textit{rewiring} algorithm is employed, initially proposed by \citet{BMST97} and \citet{WS98}. 

\begin{figure}[htp]
\begin{center}
\includegraphics[width=0.20\columnwidth]{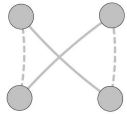} 
\caption{Example of a rewired network. Dashed edges are rewired, i.e. replaced by solid edges.}
\label{fig:rewiring}
\end{center}
\end{figure}

The algorithm iteratively rewires edges in some component $c$, meaning that we randomly choose two edges in $E_c$, $\{v_i,v_j\}$ and $\{v_k,v_l\}$, and switch one of theirs incident vertices. The resulting edges are e.g. $\{v_i,v_l\}$ and $\{v_k,v_j\}$ (see \figref{fig:rewiring}). 
The number of vertices and edges does not change during the rewiring process and the values for some $H_i$ can thus be assessed by generating a sufficient number of such random networks (for each component).

The details of the rewiring algorithm are omitted due to space limitations, we only discuss two aspects. First, the number of rewirings should be kept relatively small (e.g. $<|E_c|$), otherwise the constructed networks are completely random with no trace of the one we start with \textendash\space (probably) not networks representing a set of collisions. 
We also want to compare components to other random ones, which are similar to them, at least in the aspect of this rewirings. If a component significantly differs even from these similar ones, there is probably some severe anomaly in it.

Second, one can notice that the algorithm never changes the degrees of the vertices. As we wish to assess the degrees as well, the algorithm can be simply adopted to the task in an \textit{ad~hoc} fashion. We add an extra vertex $v_e$ and connect all other vertices with it. As this vertex is removed at the end, rewiring one of the newly added edges with some other (old) edge changes the degrees of the vertices. Let $\{v_i,v_e\}$, $\{v_k,v_l\}$ be the edges being rewired and let $\{v_i,v_l\}$, $\{v_k,v_e\}$ be the edges after the rewiring. The (true) degree of vertex $v_i, v_k$ was increased, decreased by one respectively.

To assess the values of indicators we separately construct random components for each component $c\in C(G)$ and indicator $I_i\in I$, and approximate the distributions for characteristics $H_i$ ($H_i$ are seen as random variables). A statistical test is employed to test the \textit{null hypothesis}, if the observed value $H_i(c)$ comes from the distribution for $H_i$. The test can be \textit{one} or \textit{two-tailed}, based on the nature of characteristic $H_i$. In the case of one-tailed test, where large values of $H_i$ are suspicious, we get
\begin{eqnarray}
I_i(c) & = & \left\{\begin{array}{cl}
1 & \hat{P}_c(H_i\geq H_i(c))<t_i \\
0 & \mbox{otherwise}
\end{array}\right.,
\label{eq_I_one_sided}
\end{eqnarray}
where \textit{probability density function} $P(H_i)$ is approximated with the generated distribution $\hat{P}_c(H_i)$ and $t_i$ is a \textit{critical threshold} or acceptable \textit{Type I error} (e.g. set to $0.05$). In the case of two-tailed test the equation~(\ref{eq_I_one_sided}) rewrites to
\begin{eqnarray}
I_i(c) & = & \left\{\begin{array}{cl}
1 & \begin{array}{l}
\hat{P}_c(H_i\geq H_i(c))<t_i/2\mbox{ }\vee \\
\mbox{ }\hat{P}_c(H_i\leq H_i(c))<t_i/2
\end{array} \\
0 & \begin{array}{l}
\mbox{otherwise}
\end{array}
\end{array}\right..
\label{eq_I_two_sided}
\end{eqnarray}

Knowing the values for all indicators $I_i$ we can now indicate the suspicious components in $C(G)$. The simplest way to accomplish this is to use a \textit{majority classifier} or \textit{voter}, indicating all the components, for which at least half of the indicators is set to $1$, as suspicious. Let $S(G)$ be a set of suspicious components in a network $G$, $S(G)\subseteq C(G)$, then
\begin{eqnarray}
S(G) & = & \{c|\mbox{ }c\in C(G)\wedge\sum_{i=1}^nI_i(c)\geq n/2\}.
\label{eq_S_majority}
\end{eqnarray}
When fraudulent components share most of the characteristics, measured by the indicators, we would clearly indicate them (they would have most, at least half, of the indicators set). Still, the approach is rather naive having three major weaknesses (among others). (1) there is no guarantee that the threshold $n/2$ is the best choice; (2) we do not consider how many components have some particular indicator set; and (3) all indicators are treated as equally important. Normally, we would use some sort of supervised learning technique that eliminates this weaknesses (e.g. regression, neural networks, classification trees, etc.), but again, due to the lack of labeled data and skewed class distribution in the collisions domain, this would only rarely be feasible (the size of $C(G)$ is even much smaller then the size of the actual data set).

To cope with the last two weaknesses mentioned, we suggest using \textit{principal component analysis of RIDITs} (\textit{PRIDIT}) proposed by \citet{BL77} (see \citep{Bro81}), which has already been used for detecting fraudulent insurance claim files \citep{BDGLA02}, but not for detecting groups of fraudsters (i.e. fraudulent components). The \textit{RIDIT} analysis was first introduced by \citet{Bro58}. 

\textit{RIDIT} is basically a scoring method that transforms a set of \textit{categorical} attribute values into a set of values from interval $[-1,1]$, thus they reflect the probability of an occurrence of some particular categorical value. Hence, an \textit{ordinal scale} attribute is transformed into an \textit{interval scale} attribute. In our case, all $I_i$ are simple binary attributes, and the \textit{RIDIT} scores, denoted $R_i$, are then just
\begin{eqnarray}
R_i(c) & = & \left\{\begin{array}{cl}
\hat{p}_i^0 & I_i(c)=1 \\
-\hat{p}_i^1 & I_i(c)=0
\end{array}\right.,
\label{eq_R_i}
\end{eqnarray}
where $c\in C(G)$, $\hat{p}_i^1$ is the \textit{relative frequency} of $I_i$ being equal to $1$, computed from the entire data set, and $\hat{p}_i^0=1-\hat{p}_i^1$.

We demonstrate the technique with an example. Let $\hat{p}_i^1$ be equal to $0.95$ \textendash\space almost all of the components have the indicator $I_i$ set. The \textit{RIDIT} score for some component $c$, with $I_i(c)=1$, is then just $0.05$, as the indicator clearly gives a poor indication of fraudulent components. On the other hand, for some component $c$, with $I_i(c)=0$, the \textit{RIDIT} score is $-0.95$, since the indicator very likely gives a good indication of the non-fraudulent components. Similar intuitive explanation can be made by setting $\hat{p}_i^1$ to $0.05$. A full discussion of \textit{RIDIT} scoring is omitted, for more details see \citep{Bro81,BL77}.

Introduction of \textit{RIDIT} scoring diminishes previously mentioned second weakness. To also cope with the third, we make use of the \textit{PRIDIT} technique. The intuition of this technique is that we can weight indicators in some ensemble by assessing the agreement of some particular indicator with the entire ensemble. We make a (probably incorrect) assumption that indicators are independent.

Formally, let $W$ be a vector of \textit{weights} for the ensemble of \textit{RIDIT scorers} $R_i$ for indicators $I_i$, denoted $W=[w_1,w_2\dots w_n]^T$, and let $R$ be a matrix with $i,j^{th}$ component equal to $R_{j}(c)$, where $c$ is an $i^{th}$ component in $C(G)$. Matrix product $RW$ gives the ensemble's score for all the components, i.e. $i^{th}$ component in vector $RW$ is equal to the weighted linear combination of \textit{RIDIT} scores for $i^{th}$ component in $C(G)$. Denote $S=RW$, we can then assess indicators agreement with entire ensemble as (written in matrix form) 
\begin{eqnarray}
 & & R^TS/\parallel R^TS\parallel.
\label{eq_W_0}
\end{eqnarray}
Equation~(\ref{eq_W_0}) computes normalized scalar products of columns of $R$, which corresponds to the returned values of \textit{RIDIT} scorers, and $S$, which is the overall score of the entire ensemble (for each component in $C(G)$). When the returned values of some scorer are completely orthogonal to the ensemble's scores, the resulting normalized scalar product equals $0$, and reaches its maximum, or minimum, when they are perfectly aligned.

Equation~(\ref{eq_W_0}) thus gives scorers (indicators) agreement with the ensemble and can be used to assign new weights, i.e. $W^1=R^TS/ ||R^TS||$. Greater weights are assigned to the scorers that kind of agree with the general belief of the ensemble. Denote $S^1=RW^1$, then $S^1$ is a vector of overall scores using these newly determined weights. There is of course no reason to stop the process here, as we can iteratively get even better weights. We can write
\begin{eqnarray}
W^i & = & \frac{R^TS^{i-1}}{||R^TS^{i-1}||} = \frac{R^TRW^{i-1}}{||R^TRW^{i-1}||}
\label{eq_W_i}
\end{eqnarray} 
for $i\geq 1$, which can be used to iteratively compute better and better weights for an ensemble of \textit{RIDIT} scorers $R_i$, starting with some weights, e.g. $W^0=[1,1\dots 1]$ \textendash\space the process converges to some fixed point no matter the starting weights (due to some assumptions). It can be shown that the fixed point is actually the \textit{first principal component} of the matrix $R^TR$ denoted $W^\infty$. For more details on \textit{PRIDIT} technique see \citep{BDGLA02}.

We can now score each component in $C(G)$ using the \textit{PRIDIT} technique for indicators $I_i$ and output as suspicious all the components, with a score greater than $0$. Thus
\begin{eqnarray}
S(G) & = & \{c|\mbox{ }c\in C(G)\wedge R(c)W^\infty\geq 0\},
\label{eq_S_pridit}
\end{eqnarray}
where $R(c)$ is a row of matrix $R$, that corresponds to component $c$. Again there is no guarantee, that the threshold $0$ is the best choice. Still, if we know the expected proportion of fraudulent components in the data set (or e.g. expected number of fraudulent collisions), we can first rank the components using \textit{PRIDIT} technique and then output only the appropriate proportion of most highly ranked components.

\subsection{Suspicious entities detection}
\label{system_entities}
In the third module of the system key entities are detected inside each previously identified suspicious component. We focus on identifying key participants, that can be later used for the identification of other key entities (collisions, vehicles, etc.). Key participants are identified by employing \textit{Iterative Assessment Algorithm (IAA)} that uses intrinsic and relational attributes of the entities. The algorithm assigns a \textit{suspicion score} to each participant, which corresponds to the likelihood of it being fraudulent.

In classical approaches over flat data, entities are assessed using only their intrinsic attributes, thus they are assessed in complete \textit{isolation} to other entities. It has been empirically shown that the \textit{assessment} can be improved by also considering the related entities, more precisely, by considering the assessment of the related entities \citep{CDI98,DR01,LG03b,LG03a,NJ00}. The assessment of an entity is \textit{inferred} from the assessments of the related entities and \textit{propagated} onward. Still, incorporating only the intrinsic attributes of the related entities generally doesn't improve, or even deteriorates, the assessment \citep{CDI98,OML00}. 

The proposed \textit{IAA} algorithm thus assesses the entities by also considering the assessment of their related entities. As these related entities were also assessed using the assessments of their related entities, and so on, the entire network is used in the assessment of some particular entity. This could not be achieved otherwise, as the formulation would surely be too complex. We proceed by introducing \textit{IAA} in a general form.

Let $c$ be some suspicious component in network $G$, $c\in S(G)$, and let $v_i$ be one of its vertices, $v_i\in V_c$. Furthermore, let $N(v_i)$ be a set of neighbor vertices of $v_i$ (i.e. vertices at distance $1$) and $V(v_i)=N(v_i)\cup\{v_i\}$, and let $E(v_i)$ be a set of edges incident to $v_i$ (i.e. $E(v_i)=\{e|\mbox{ }e\in E_c\wedge v_i\in e\}$). Let also $en_i$ be an entity corresponding to vertex $v_i$ and $N(en_i)$, $V(en_i)$ be a set of entities that corresponds to $N(v_i)$, $V(v_i)$ respectively. We define the suspicion score $s$, $s(\cdot)\geq 0$, for the entity $en_i$ as
\begin{eqnarray}
\label{eq_assess}
s(en_i) & = & AM(s(N(en_i)),V(en_i),V(v_i),E(v_i)) \\
 & = & AM(i,c),
\nonumber
\end{eqnarray}
where $AM$ is some \textit{assessment model} and $s(N(en_i))$ is a set of suspicion scores for entities in $N(en_i)$. The suspicion of some entity is dependent on the assessment of the related entities (first argument in equation~(\ref{eq_assess})), on the intrinsic attributes of related entities and itself (second argument), and on the relational attributes of the entity (last two arguments). We assume that $AM$ is \textit{linear} in the assessments of the related entities (i.e. $s(N(en_i))$) and that it returns higher values for fraudulent entities.

For some entity $en_i$, when the suspicion scores of the related entities are known, $en_i$ can be assessed using equation~(\ref{eq_assess}). Commonly, none of the suspicion scores are known preliminary (as the data set is unlabeled), and the equation thus cannot be used in a common manner. Still, one can incrementally assess the entities in an \textit{iterative} fashion, similar to e.g. \citep{BP98,Kle99}.

Let $s^{0}(\cdot)$ be some set of suspicion scores, e.g. $s^{0}(\cdot)=1$. We can then assess the entities using scores $s^{0}(\cdot)$ and equation~(\ref{eq_assess}), and get better scores $s^{1}(\cdot)$. We proceed with this process, iteratively refining the scores until some stopping criteria is reached. Generally, on the $k^{th}$ iteration, entities are assessed using
\begin{eqnarray}
\label{eq_assess_k}
s^k(en_i) & = & AM(s^{k-1}(N(en_i)),V(en_i),V(v_i),E(v_i)) \\
 & = & AM(i,k,c).
\nonumber
\end{eqnarray}
Note that the choice for $s^0(\cdot)$ is arbitrary \textendash\space the process converges to some \textit{fixed point} no matter the starting scores (due to some assumptions). Hence, the entities are assessed without preliminary knowing any suspicion score to bootstrap the procedure.

We present the \textit{IAA} algorithm below.
\begin{table}[ht]
\begin{center}
\begin{tabular}{|lllll}
\multicolumn{5}{l}{\textit{IAA algorithm}} \\\hline\hline
& \multicolumn{4}{l|}{$s^{0}(\cdot)=1$} \\
& \multicolumn{4}{l|}{$k=1$} \\
& \multicolumn{4}{l|}{\texttt{WHILE NOT} \textit{stopping criteria} \texttt{DO}} \\
& & \multicolumn{3}{l|}{\texttt{FOR} $\forall v_i,en_i$ \texttt{DO}} \\
& & & \multicolumn{2}{l|}{$s^k(en_i) = \alpha s^{k-1}(en_i) + (1-\alpha) AM(i,k,c)$} \\
& & \multicolumn{3}{l|}{\texttt{FOR} $\forall v_i,en_i$: $v_i$ \textit{non-bucket} \texttt{DO}} \\
& & & \multicolumn{2}{l|}{\textit{normalize} $s^k(en_i)$} \\
& & \multicolumn{3}{l|}{$k=k+1$} \\
& \multicolumn{4}{l|}{\texttt{RETURN }$s^k(\cdot)$} \\\hline
\end{tabular}
\end{center}
\end{table}

Entities are iteratively assessed using model $AM$ ($\alpha$ is a \textit{smoothing parameter} set to e.g. $0.75$). In order for the process to converge, scores corresponding to \textit{non-bucket} vertices are normalized at the end of each iteration. Due to the fact that relations represented by the networks are often not binary, there are usually some vertices only serving as \textit{buckets} that store the suspicion assessed at this iteration to be propagated on the next. \textit{Non-bucket} vertices correspond to entities that are actually being assessed and only these scores should be normalized (for binary relations all the vertices are of this kind). Structure of such \textit{bucket} networks would typically correspond to \textit{bipartite graphs}\footnote{In the social science literature bipartite graphs are known as \textit{collaboration networks}.} \textendash\space bucket vertices would only be connected to non-bucket vertices (and vice versa). In the case of \textit{COPTA} networks, used in this module of the (prototype) system, bucket vertices are those representing collisions. 

One would intuitively run the algorithm until some fixed point is reached, i.e. when the scores no longer change. We empirically show that, despite the fact that iterative assessment does indeed increase the performance, such approach actually decreases it. The reason is that the scores \textit{over-fit} the model. We also show, that superior performance can be achieved with a dynamic approach \textendash\space by running the algorithm for $d(c)$ iterations (diameter of component $c$). For more see sections~\ref{evaluation},~\ref{discussion}.

Note that if each subsequent iteration of the algorithm actually increased the performance, one could assess the entities directly. When $AM$ is linear in the assessments of related entities, the model could be written as a set of \textit{linear equations} and solved exactly (analytically).

An arbitrary model can be used with the algorithm. We propose several linear models based on the observation that in many of these bucket networks the following holds: \textit{every entity is well defined with (only) the entities directly connected to it, considering the context observed}. E.g. in the case of \textit{COPTA} networks, every collision is connected to its participants, who are clearly the ones who ``define'' the collision, and every participant is connected with its collisions, which are the precise aspect of the participant we wish to investigate when dealing with fraud detection. Similar discussion could be made for movie-actor, corporate board-director and other well known collaboration networks. A model using no attributes of the entities is thus simply the sum of suspicion scores of the related entities (we omit the arguments of the model)
\begin{eqnarray}
AM_{raw} & = & \sum_{\{v_i,v_j\}\in E(v_i)} s(en_j).
\label{eq_model_raw}
\end{eqnarray}
Our empirical evaluation shows that even such a simple model can achieve satisfactory performance.

To incorporate entities' attributes into the model, we introduce \textit{factors}. These are based on intrinsic or relational attributes of entities. The intuition behind the first is that some intrinsic attributes' values are highly correlated with fraudulent activity. Suspicion scores of corresponding entities should in this case be increased and also propagated on the related entities. Moreover, many of the relational attributes (i.e. labels of the edges) increase the likelihood of fraudulent activity \textendash\space the propagation of suspicion over such edges should also be increased.

Let $l_{E_G}$ be the edge labeling function and $\Sigma_{E_G}$ the alphabet of all possible edge labels, i.e. $\Sigma_{E_G}=\{Driver,Passenger\dots\}$ (for \textit{COPTA} networks). Furthermore, let $En$ be a set of all entities $en_i$. We define $F_{int}$, $F_{rel}$ to be the factors, corresponding to intrinsic, relational attributes respectively, as
\begin{eqnarray}
F_{int}: & En \rightarrow [0,\infty), \\
F_{rel}: & \Sigma_{E_G}\rightarrow [0,\infty).
\end{eqnarray}
Improved model incorporating these factors is then
\begin{eqnarray}
AM_{bas} & = & F_{int}(en_i)\sum_{e=\{v_i,v_j\}\in E(v_i)} F_{rel}(l_{E_G}(e))\mbox{ }s(en_j).
\label{eq_model_basic}
\end{eqnarray}
Factors $F_{int}$ are computed from (similar for $F_{rel}$)
\begin{eqnarray}
F_{int}(en_i) & = & \prod_k F_{int}^k(en_i)
\label{eq_F_int}
\end{eqnarray}
where
\begin{eqnarray}
F_{int}^k(en_i) & = & \left\{\begin{array}{cl}
1/(1-f_{int}^k(en_i)) & f_{int}^k(en_i)\geq 0 \\
1+f_{int}^k(en_i) & \mbox{otherwise}
\end{array}\right.
\label{eq_f_int}
\end{eqnarray}
and
\begin{eqnarray}
f_{int}^k: & En\rightarrow (-1,1).
\end{eqnarray}
$f_{int}^k$ are \textit{virtual factors} defined by the domain expert. The transformation with equation~(\ref{eq_f_int}) is done only to define factors on the interval $(-1,1)$, rather than on $[0,\infty)$. The first is more intuitive as e.g. two ``opposite'' factors are now $f$ and $-f$, $f\in[0,1)$, opposed to $f$ and $1/f$, $f>0$, before.

Some virtual factor $f_{int}^k$ can be an arbitrary function defined due to a single attribute of some entity, or due to several attributes formulating \textit{correlations} between the attributes. When attributes' values correspond to some suspicious activity (e.g. collision corresponds to some classical \textit{scheme}), factors are set to be close to $1$, and close to $-1$, when values correspond to non-suspicious activity (e.g. children in the vehicle). Otherwise, they are set to be $0$.

Note that assessment of some participant with models $AM_{raw}$ and $AM_{bas}$ is highly dependent on the number of collisions this participant was involved in. More precisely, on the number of the terms in the sums in equations~(\ref{eq_model_raw}),~(\ref{eq_model_basic}) (which is exactly the degree of the corresponding vertex). Although this property is not vain, we still implicitly assume we posses \textit{all} of the collisions a certain participant was involved in. This assumption is often not true (in practice).

We propose a third model diminishing the mentioned assumption. Let $\overline{d_G}$ be the average degree of the vertices in network $G$, $\overline{d_G}=ave\{d(v_k)|\mbox{ }v_k\in V_G\}$. The model is
\begin{eqnarray}
AM_{\cdot}^{mean} & = & \frac{\overline{d_G}+d(v_i)}{2}\frac{AM_\cdot}{d(v_i)} = \left(1+\frac{\overline{d_G}}{d(v_i)}\right)\frac{AM_\cdot}{2},
\label{eq_model_laplace}
\end{eqnarray}
where $AM_{\cdot}$ can be any of the models $AM_{raw}$, $AM_{bas}$. $AM_{\cdot}^{mean}$ averages terms in the sum of the model $AM_{\cdot}$, and multiplies this average by the mean of vertex's degree and the average degree over all the vertices in $V_G$. Thus a sort of \textit{Laplace smoothing} is employed that pulls the vertex degree toward the average, in order to diminish the importance of this parameter in the final assessment. Empirical analysis in section~\ref{evaluation} shows that such a model outperforms the other two.

Knowing scores $s(\cdot)$ for all the entities in some connected component $c\in G$, one can rank them according to the suspicion of their being fraudulent. In order to also compare the entities from various components, scores must be normalized appropriately (e.g. multiplied with the number of collisions represented by component $c$).

\subsection{Final remarks}
\label{system_remarks}
In the previous section (third module of the system) we focused only on detection of fraudulent participants. Their suspicion scores can now be used for assessment of other entities (e.g. collisions, vehicles), using one of the assessment models proposed in section~\ref{system_entities}.

When all of the most highly ranked participants in some suspicious component are directly connected to each other (or through buckets), they are proclaimed to belong to the same group of fraudsters. Otherwise they belong to several groups. During the investigation process, the domain expert or investigator analyzes these groups and determines further actions for resolving potential fraud. Entities are investigated in the order induced by scores $s(\cdot)$.

\begin{figure}[htp]
\begin{center}
\includegraphics[width=1.00\columnwidth]{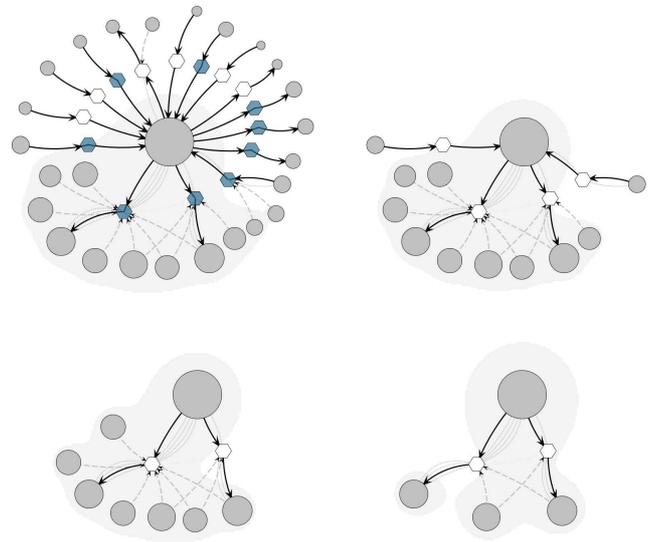} 
\caption{Four \textit{COPTA} networks showing same group of collisions. Size of the participants' vertices correspond to their suspicion score; only participants with score above some threshold, and connecting collisions, are shown on each network. The contour was drawn based on the \textit{harmonic mean} distance to every vertex, weighted by the suspicion scores. (Blue) filled collisions' vertices in the first network correspond to collisions that happened at night.}
\label{fig:visualization}
\end{center}
\end{figure}

Networks also allow a neat visualization of the assessment (see \figref{fig:visualization}).

\section{Evaluation with the prototype system}
\label{evaluation}
We implemented a prototype system to empirically evaluate the performance of the proposition. Furthermore, various components of the system are analyzed and compared to other approaches. To simulate realistic conditions, the data set used for evaluation consisted only of the data, that can be easily (automatically) retrieved from police records (\textit{semistructured data}). We report results of the assessment of participants (not e.g. collisions).

\subsection{Data}
\label{data}
The data set consists of $3451$ participants involved in $1561$ collisions in Slovenia between the years $1999$ and $2008$. The set was made by merging two data sets, one labeled and one unlabeled.

The first, labeled, consists of collisions corresponding to previously identified fraudsters and some other participants, which were investigated in the past. In a few cases, when \textit{class} of a participant could not be determined, it was set according to the domain expert's and investigator's belief. As the purpose of our system is to identify groups of fraudsters, and not some isolated fraudulent collisions, (almost) all isolated collisions were removed from this set. It is thus a bit smaller (i.e. $211$ participants, $91$ collisions), but still large enough to make the assessment.

To achieve a more realistic class distribution and better statistics for \textit{PRIDIT} analysis, the second larger data set was merged with the first. The set consists of various collisions chosen (almost) at random, although some of them are still related with others. Since random data sampling is not advised for relational data \citep{Jen99}, this set is used explicitly for \textit{PRIDIT} analysis. Both data sets consist of only standard collisions (e.g. there are no chain collisions involving numerous vehicles or coaches with many passengers).

Class distribution for the data set can be seen in table~\ref{tbl:class_distribution}.

\begin{table}[htp]
\begin{center}
\begin{tabular}{cccc} 
\multicolumn{4}{l}{\textit{Class distribution}} \\\hline\hline
& \textit{Count} & \multicolumn{2}{c}{\textit{Proportion}} \\\hline
Fraudster & $46$ & $1.3 \%$ & $21.8 \%$  \\
Non-fraudster & $165$ & $4.8 \%$ & $78.2 \%$  \\
Unlabeled & $3240$ & $93.9 \%$ &  \\
\end{tabular}
\caption{Class distribution for the data set used in the analysis of the proposed expert system.}
\label{tbl:class_distribution}
\end{center}
\end{table}

The entire assessment was made using the merged data set, while the reported results naturally only correspond to the first (labeled) set. Note that the assessment of entities in some connected component is completely independent of the entities in other components (except for \textit{PRIDIT} analysis).

\subsection{Results}
\label{results}
Performance of the system depends on random generation of networks, used for detection of suspicious components (second module). We construct $200$ random networks for each indicator and each component (equations~(\ref{eq_I_one_sided}),~(\ref{eq_I_two_sided})), however, the results still vary a little. The entire assessment was thus repeated $20$ times and the scores were averaged. To assess the ranking of the system, average $AUC$ (\textit{Area Under Curve}) scores were computed, $\overline{AUC}$. Results given in tables~\ref{tbl:assessment_models},~\ref{tbl:factors},~\ref{tbl:iaa_algorithm},~\ref{tbl:fraudulent_components} are all $\overline{AUC}$. 

\begin{table}[ht]
\begin{center}
\begin{tabular}{clc} 
\multicolumn{3}{l}{\textit{Golden standard}} \\\hline\hline
$CA$ & $0.8720$ & \\
\textit{Recall} & $0.8913$ & \\
\textit{Precision} & $0.6508$ & \\
\textit{Specificity} & $0.8667$ & \\
$F1$ \textit{score} & $0.7523$ & \\
$\overline{AUC}$ & $\mathbf{0.9228}$ & \\
\end{tabular}
\caption{Performance of the system that uses \textit{PRIDIT} analysis with $IAA^{mean}_{bas}$ algorithm. Various metrics are reported; all except $\overline{AUC}$ are computed so the total cost (on the first run) is minimal.}
\label{tbl:golden_standard}
\end{center}
\end{table}

In order to obtain a standard for other analyses, we first report the performance of the system that uses \textit{PRIDIT} analysis for fraudulent components detection, and \textit{IAA} algorithm with model $AM_{bas}^{mean}$ for fraudulent entities detection, denoted $IAA^{mean}_{bas}$ (see table~\ref{tbl:golden_standard}). Various metrics are computed, i.e. \textit{classification accuracy} ($CA$), \textit{recall} (\textit{true positive rate}), \textit{precision} (\textit{positive predictive value}), \textit{specificity} ($1-$ \textit{false positive rate}), \textit{$F1$ score} (\textit{harmonic mean of recall and precision}) and $\overline{AUC}$. All but last are metrics that assess the classification of some approach, thus a threshold for suspicion scores must be defined. We report the results from the first run that minimize the total cost, assuming the cost of misclassified fraudsters and non-fraudsters is the same. Same holds for confusion matrix seen in table~\ref{tbl:confusion_matrix}.

\begin{table}[ht]
\begin{center}
\begin{tabular}{ccc} 
\multicolumn{3}{l}{\textit{Confusion matrix}} \\\hline\hline
& \textit{Suspicious} & \textit{Unsuspicious} \\\hline
Fraudster & $41$ & $5$  \\
Non-fraudster & $22$ & $143$ \\
\end{tabular}
\caption{Confusion matrix for the system that uses \textit{PRIDIT} analysis with $IAA^{mean}_{bas}$ algorithm (determined so the total cost on the first run is minimal).}
\label{tbl:confusion_matrix}
\end{center}
\end{table}

We proceed with an in-depth analysis of the proposed \textit{IAA} algorithm. Table~\ref{tbl:assessment_models} shows the results of the comparison of different assessment models, i.e. $IAA_{raw}$, $IAA_{bas}$, $IAA^{mean}_{raw}$ and $IAA^{mean}_{bas}$. Factors for models $IAA_{bas}$ and $IAA^{mean}_{bas}$ (equation~(\ref{eq_f_int})) were set by the domain expert, with the help of statistical analysis of data from collisions. To further analyze the impact of factors on final assessment, an additional set of factors was defined by the authors. Values were set due to authors' intuition; corresponding models are $IAA_{int}$ and $IAA^{mean}_{int}$. Results of the analysis can be seen in table~\ref{tbl:factors}.

\begin{table}[ht]
\begin{center}
\begin{tabular}{cccc} 
\multicolumn{4}{l}{\textit{Assessment models}} \\\hline\hline
\multicolumn{4}{c}{\textit{PRIDIT}} \\\hline
$IAA_{raw}$ & $IAA_{bas}$ & $IAA^{mean}_{raw}$ & $IAA^{mean}_{bas}$ \\\hline
$0.8872$ & $0.9145$ & $0.8942$ & $0.9228$ \\
\end{tabular}
\caption{Comparison of different assessment models for \textit{IAA} algorithm (after \textit{PRIDIT} analysis).}
\label{tbl:assessment_models}
\end{center}
\end{table}

\begin{table}[ht]
\begin{center}
\begin{tabular}{ccc} 
\multicolumn{3}{l}{\textit{Factors}} \\\hline\hline
\multicolumn{3}{c}{\textit{ALL}} \\\hline
$IAA^{mean}_{raw}$ & $IAA^{mean}_{int}$ & $IAA^{mean}_{bas}$ \\\hline
$0.8188$ & $0.8435$ & $0.8787$ \\
\\\hline\hline
\multicolumn{3}{c}{\textit{PRIDIT}} \\\hline
$IAA^{mean}_{raw}$ & $IAA^{mean}_{int}$ & $IAA^{mean}_{bas}$ \\\hline
$0.8942$ & $0.9086$ & $0.9228$ \\
\end{tabular}
\caption{Analysis of the impact of factors on the final assessment (on all the components and after \textit{PRIDIT} analysis).}
\label{tbl:factors}
\end{center}
\end{table}

\begin{figure}[htp]
\begin{center}
\includegraphics[width=1.00\columnwidth]{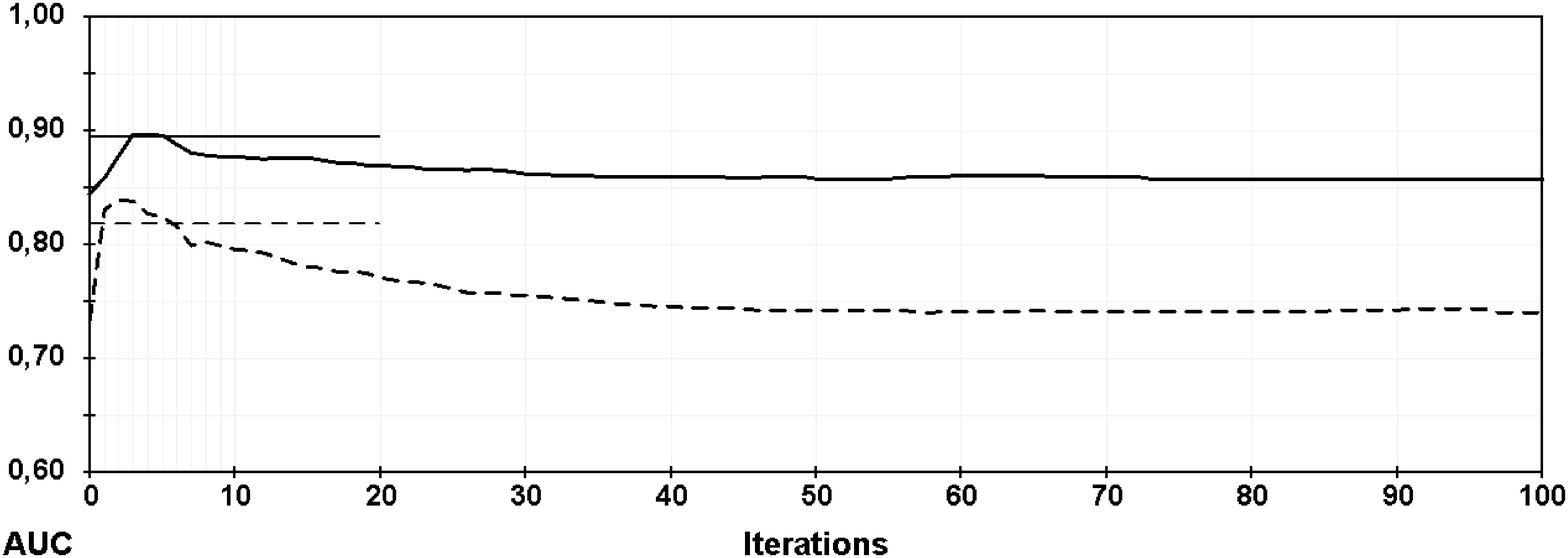} 
\caption{$AUC$ scores with respect to the number of iterations made in the \textit{IAA} algorithm. Solid curves correspond to $IAA^{mean}_{raw}$ algorithm after \textit{PRIDIT} analysis and dashed curves to $IAA^{mean}_{raw}$ algorithm ran on all the components. Straight line segments show the scores achieved with dynamic setting of the number of iterations (see text).}
\label{fig:iterations_raw}
\end{center}
\end{figure}

\begin{figure}[htp]
\begin{center}
\includegraphics[width=1.00\columnwidth]{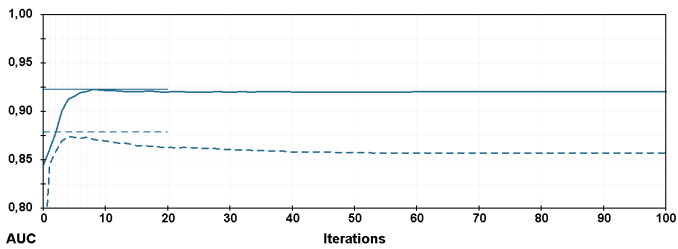} 
\caption{$AUC$ scores with respect to the number of iterations made in the \textit{IAA} algorithm. Solid curves correspond to $IAA^{mean}_{bas}$ algorithm after \textit{PRIDIT} analysis and dashed curves to $IAA^{mean}_{bas}$ algorithm ran on all the components. Straight line segments show the scores achieved with dynamic setting of the number of iterations (see text).}
\label{fig:iterations_bas}
\end{center}
\end{figure}

As already mentioned, the performance of the \textit{IAA} algorithm depends on the number of iterations made in the assessment (see section~\ref{system_entities}). We have thus plotted the $AUC$ scores with respect to the number of iterations made (for the first run), in order to clearly see the dependence; plots for $IAA^{mean}_{raw}$, $IAA^{mean}_{bas}$ can be seen in \figref{fig:iterations_raw}, \figref{fig:iterations_bas} respectively. We also show that superior performance can be achieved, if the number of iterations is set dynamically. More precisely, the number of iterations made for some component $c\in C(G)$ is
\begin{eqnarray}
 & max\{\overline{d_G},d(c)\},
\label{eq_dyn_iters}
\end{eqnarray}
where $d(c)$ is the diameter of $c$ and $\overline{d_G}$ the average diameter over all the components. All other results reported in this analysis used such a dynamic setting.

Due to the lack of other expert system approaches for detecting groups of fraudsters, or even individual fraudsters (according to our knowledge), no comparative analysis of such kind could be made. The proposed \textit{IAA} algorithm is thus compared against several well known measures for anomaly detection in networks \textendash\space \textit{betweenness centrality} (\textit{BetCen}), \textit{closeness centrality} (\textit{CloCen}), \textit{distance centrality} (\textit{DisCen}) and \textit{eigenvector centrality} (\textit{EigCen}) \citep{Fre77,Fre79}. They are defined as
\begin{eqnarray}
BetCen(v_i) & = &  \sum_{v_j,v_k\in V_c} \frac{g_{v_j,v_k}(v_i)}{g_{v_j,v_k}},\\
\label{eq_bet_cen}
CloCen(v_i) & = & \frac{1}{n_c-1}\sum_{v_j\in V_c\backslash{v_i}}d(v_i,v_j) ,\\
DegCen(v_i) & = & \frac{d(v_i)}{n_c-1} ,\\
EigCen(v_i) & = &  \frac{1}{\lambda} \sum_{\{v_i,v_j\}\in E_c} EigCen(v_j) ,
\end{eqnarray}
where $n_c$ is the number of vertices in component $c$, $n_c=|V_c|$, $\lambda$ is a constant, $g_{v_j,v_k}$ is the number of geodesics between vertices $v_j$ and $v_k$ and $g_{v_j,v_k}(v_i)$ the number of such geodesics that pass through vertex $v_i$, $i\neq j\neq k$. For further discussion see \citep{Fre77,Fre79,New03}. 

These measures of centrality were used to assign suspicion score to each participant; scores were also appropriately normalized as in the case of \textit{IAA} algorithm. For a fair comparison, measures were compared against the model that uses no intrinsic attributes of entities, i.e. $IAA^{mean}_{raw}$. The results of the analysis are shown in table~\ref{tbl:iaa_algorithm}.

\begin{table}[ht]
\begin{center}
\begin{tabular}{ccccc} 
\multicolumn{5}{l}{\textit{IAA algorithm}} \\\hline\hline
\multicolumn{5}{c}{\textit{ALL}} \\\hline
\textit{BetCen} & \textit{CloCen} & \textit{DegCen} & \textit{EigCen} & $IAA^{mean}_{raw}$\\\hline
$0.6401$ & $0.8138$ & $0.7428$ & $0.7300$ & $0.8188$ \\
\\\hline\hline
\multicolumn{5}{c}{\textit{PRIDIT}} \\\hline
\textit{BetCen} & \textit{CloCen} & \textit{DegCen} & \textit{EigCen} & $IAA^{mean}_{raw}$\\\hline
$0.6541$ & $0.8158$ & $0.8597$ & $0.8581$ & $0.8942$ \\
\end{tabular}
\caption{Comparison of the \textit{IAA} algorithm against several well known measures for anomaly detection in networks (on all the components and after \textit{PRIDIT} analysis). For a fair comparison, no intrinsic attributes are used in the \textit{IAA} algorithm (i.e. model $AM^{mean}_{raw}$).}
\label{tbl:iaa_algorithm}
\end{center}
\end{table}

Next, we analyzed different approaches for detection of fraudulent components (see table~\ref{tbl:fraudulent_components}). The same set of $9$ indicators was used for the majority voter (equation~(\ref{eq_S_majority})) and for \textit{(P)RIDIT} analysis (equation~(\ref{eq_S_pridit})). For the latter, we use a variant of \textit{random undersampling} (\textit{RUS}), to cope with skewed class distribution. We output most highly ranked components, thus the set of selected components contain $4\%$ of all the collisions (in the merged data set) Analyses of automobile insurance fraud mainly agree that up to $20\%$ of all the collisions are fraudulent, and up to $20\%$ of the latter correspond to non-opportunistic fraud (various resources). However, for the majority voter, such an approach actually decreases performance \textendash\space we therefore report results where all components, with at least half of the indicators set, are selected.

Several individual indicators, achieving superior performance, are also reported. Indicator $I_{BetCen}$ is based on betweenness centrality (equation~(\ref{eq_bet_cen})), $I_{MinCov}$ on minimum vertex cover and $I_{l^{-1}}$ on $l^{-1}$\textit{ measure} defined as the harmonic mean distance between every pair of vertices in some component $c$,
\begin{eqnarray}
l^{-1} & = & \frac{1}{\frac{1}{2}n_c(n_c+1)}\sum_{v_i,v_j\in V_c, i\geq j}d(v_i,v_j)^{-1}.
\end{eqnarray}

\begin{table}[ht]
\begin{center}
\begin{tabular}{cccccc} 
\multicolumn{6}{l}{\textit{Fraudulent components}} \\\hline\hline
$I_{MinCov}$ & $I_{l^{-1}}$ & $I_{BetCen}$ & \textit{MAJOR} & \textit{RIDIT} & \textit{PRIDIT} \\\hline
\multicolumn{6}{c}{\textit{ALL}} \\\hline
$0.6019$ & $0.6386$ & $0.6774$ & $0.7946$ & $0.6843$ & $0.7114$ \\
\\\hline\hline
$I_{MinCov}$ & $I_{l^{-1}}$ & $I_{BetCen}$ & \textit{MAJOR} & \textit{RIDIT} & \textit{PRIDIT} \\\hline
\multicolumn{6}{c}{$IAA^{mean}_{bas}$} \\\hline
$0.6119$ & $0.8494$ & $0.8549$ & $0.8507$ & $0.9221$ & $0.9228$ \\
\end{tabular}
\caption{Comparison of different approaches for detection of fraudulent components (prior to no fraudulent entities detection and $IAA^{mean}_{bas}$).}
\label{tbl:fraudulent_components}
\end{center}
\end{table}

We last analyze the importance of proper data representation for detection of groups of fraudsters \textendash\space the use of networks. Networks were thus transformed into flat data and some standard unsupervised learning techniques were examined (e.q. \textit{k-means}, \textit{hierarchical clustering}). We obtained no results comparable to those given in table~\ref{tbl:golden_standard}.

Furthermore, we tested nine standard supervised data-mining techniques to analyze the compensation of data labels for the inappropriate representation of data. We used (default) implementations of classifiers in \textit{Orange} data-mining software \citep{DZLC04} and $20$-\textit{fold} \textit{cross validation} was employed as the validation technique. Best performance, up to $AUC\approx 0.86$, was achieved with \textit{Naive Bayes}, \textit{support vector machines}, \textit{random forest} and, interestingly, also \textit{k-nearest neighbors} classifier. Scores for other approaches were below $AUC=0.80$ (e.g. \textit{logistic regression}, \textit{classification trees}, etc.).

\section{Discussion}
\label{discussion}
Empirical evaluation from the previous section shows that automobile insurance fraud can be detected using the proposition. Moreover, the results suggest that appropriate data representation is vital \textendash\space even a simple approach over networks can detect a great deal of fraud. The following section discusses the results in greater detail (in the order given).

Almost all of the metrics obtained with \textit{PRIDIT} analysis and $IAA^{mean}_{bas}$ algorithm, \textit{golden standard}, are very high (table~\ref{tbl:golden_standard}). Only precision appears low, still this results (only) from the skewed class distribution in the domain. The $F1$ measure is consequently also a bit lower, else the performance of the system is more than satisfactory. The latter was confirmed by the experts and investigators from a Slovenian insurance company, who were also pleased with the visual representation of the results.

The confusion matrix given in table~\ref{tbl:confusion_matrix} shows that we correctly classified almost $90\%$ of all fraudsters and over $85\%$ of non-fraudsters. Only $5$ fraudsters were not detected by the prototype system. We thus obtained a particularly high recall, which is essential for all fraud detection systems. The majority of unlabeled participants were classified as unsuspicious (not shown in table~\ref{tbl:confusion_matrix}), but the corresponding collisions are mainly isolated and the participants could have been trivially eliminated anyway (for our purposes).

We proceed with discussion of different assessment models (table~\ref{tbl:assessment_models}). Performance of the simplest of the models $IAA_{raw}$, which uses no domain expert's knowledge, could already prove sufficient in many circumstances. It can still be significantly improved by also considering the factors, set by the domain expert (model $IAA_{bas}$). Model $IAA^{mean}_{\cdot}$ further improves the assessment of both (simple) models, confirming the hypothesis behind it (see section~\ref{system_entities}). Although the models (probably incorrectly) assume that the fraudulence of an entity is linear (in the fraudulences of the related entities), they give a good approximation of the fraudulent behavior. 

The analysis of the factors used in the models confirms their importance for the final assessment. As expected, model $IAA^{mean}_{bas}$ outperforms $IAA^{mean}_{int}$, and the latter outperforms $IAA^{mean}_{raw}$ (table~\ref{tbl:factors}). First, this confirms the hypothesis that domain knowledge can be incorporated into the model using factors (as defined in section~\ref{system_entities}). Second, it shows that better understanding of the domain can improve assignment of factors. Combination of both makes the system extremely flexible, allowing for detection of new types of fraud immediately after they have been noticed by the domain expert or investigator.

As already mentioned, running the \textit{IAA} algorithm for too long over-fits the model and decreases algorithm's final performance (see \figref{fig:iterations_raw}, \figref{fig:iterations_bas}, note different scales used). Early iterations of the algorithm still increase the performance in all cases analyzed, which proves the importance of iterative assessment as opposed to \textit{single-pass} approach. However, after some particular number of iterations has been reached, performance decreases (at least slightly). Also note that the decrease is much larger in the case of $AM^{mean}_{raw}$ model than $AM^{mean}_{bas}$, indicating that the latter is superior to the first. We propose to use this \textit{decrease in performance} as an additional evaluation of any model used with \textit{IAA}, or similar, algorithm.

It is preferable to run the algorithm for only a few iterations for one more reason. Networks are often extremely large, especially when they describe many characteristics of entities. In this case, running the algorithm until some fixed point is simply not feasible. Since the prototype system uses only the basic attributes of the entities, the latter does not present a problem.

The number of iterations that achieves the best performance clearly depends on various factors (data set, model, etc.). Our evaluation shows that superior, or at least very good, performance (\figref{fig:iterations_raw}, \figref{fig:iterations_bas}) can be achieved with the use of a dynamic setting of the number of iterations (equation~(\ref{eq_dyn_iters})). 

When no detection of fraudulent components is done, the comparison between \textit{IAA} algorithm and measures of centrality shows no significant difference (table~\ref{tbl:iaa_algorithm}). On the other hand, when we use \textit{PRIDIT} analysis for fraudulent components detection, the \textit{IAA} algorithm dominates others. Still, the results obtained with \textit{DegCen} and \textit{EigCen} are comparable to those obtained with supervised approaches over flat data. This shows that even a simple approach can detect a reasonably large portion of fraud, if appropriate representation of data is used (networks).

The analysis of different approaches for detection of fraudulent components produces no major surprises (table~\ref{tbl:fraudulent_components}) \textendash\space the best results are obtained using \textit{(P)RIDIT} analysis. Note that a single indicator can match the performance of majority classifier \textit{MAJOR}, confirming its naiveness (see section~\ref{system_components}); exceptionally high $\overline{AUC}$ score obtained by \textit{MAJOR}, prior to no fraudulent entities detection, only results from the fact, that the returned set of suspicious components is almost $10$-times smaller than for other approaches. The precision of the approach is thus much higher, but for the price of lower recall (useless for fraud detection).

We have already discussed the purpose of hierarchical detection of groups of fraudsters \textendash\space to simplify detection of fraudulent entities with appropriate detection of fraudulent components. Another implication of such an approach is also simpler, or is some cases even feasible, \textit{data collection} process. As the detection of components is done using only the relations between entities (relational attributes), a large portion of data can be discarded without knowing the values of any of the intrinsic attributes. This characteristic of the system is vital when deploying in practice \textendash\space (complete) data  often cannot be obtained for all the participants, due to sensitivity of the domain.

Last, we briefly discuss the applicability of the proposition in other domains. The presented \textit{IAA} algorithm can be used for arbitrary assessment of entities over some relational domain, exploring the relations between entities with no demand for an (initial) labeled data set. When every entity is well defined with (only) the entities directly related to it, considering the context observed, one of the proposed assessment models can also be used. Furthermore, the presented framework (four modules of the system) could be employed for fraud detection in other domains. The system is also applicable for use in other domains, where we are interested in groups of related entities sharing some particular characteristics. The framework exploits the relations between entities, in order to improve the assessment, and is structured hierarchically, to make it applicable in practice.

\section{Conclusion}
\label{conclusion}
The article proposes a novel expert system approach for detection of groups of automobile insurance fraudsters with networks. Empirical evaluation shows that such fraud can be efficiently detected using the proposition and, in particular, that proper representation of data is vital. For the system to be applicable in practice, no labeled data set is used. The system rather allows the imputation of domain expert's knowledge, and it can thus be adopted to new types of fraud as soon as they are noticed. The approach can aid the domain investigator to detect and investigate fraud much faster and more efficiently. Moreover, the employed framework is easy to implement and is also applicable for detection (of fraud) in other relational domains.

Future research will be focused on further analyses of different assessment models for \textit{IAA} algorithm, considering also the nonlinear models. Moreover, the \textit{IAA} will be altered into an \textit{unsupervised algorithm}, learning the factors of the model in an unsupervised manner during the actual assessment. The factors would thus not have to be specified by the domain expert. Applications of the system in other domains will also be investigated. 

\section*{Acknowledgements}
\label{acknowledgement}
Authors thank Matja\v z Kukar, from University of Ljubljana, and Jure Leskovec, (currently) from Cornell University, for all the effort and useful suggestions; Tja\v sa Krisper Kutin for lectoring the article; and Optilab d.o.o. for the data used in the study. This work has been supported by the Slovene Research Agency \textit{ARRS} within the research program P2-0359.







\begin{thebibliography}{49}
\expandafter\ifx\csname natexlab\endcsname\relax\def\natexlab#1{#1}\fi
\expandafter\ifx\csname url\endcsname\relax
  \def\url#1{{\tt #1}}\fi
\expandafter\ifx\csname urlprefix\endcsname\relax\def\urlprefix{URL }\fi

\bibitem[{Artis et~al.(2002)Artis, Ayuso, \& Guillen}]{AAG02}
Artis, M., Ayuso, M., \& Guillen, M. (2002).
\newblock Detection of automobile insurance fraud with discrete choice models
  and misclassified claims.
\newblock {\em Journal of Risk and Insurance\/}, {\em 69\/}(3), 325--340.

\bibitem[{Ball et~al.(1997)Ball, Mollison, \& Scalia-Tomba}]{BMST97}
Ball, F., Mollison, D., \& Scalia-Tomba, G. (1997).
\newblock Epidemics with two levels of mixing.
\newblock {\em Annals of Applied Probability\/}, {\em 7\/}(1), 46--89.

\bibitem[{Barabasi \& Albert(1999)}]{BA99}
Barabasi, A.~L., \& Albert, R. (1999).
\newblock Emergence of scaling in random networks.
\newblock {\em Science\/}, {\em 286\/}(5439), 509--512.

\bibitem[{Bolton \& Hand(2002)}]{BH02}
Bolton, R.~J., \& Hand, D.~J. (2002).
\newblock Statistical fraud detection: A review.
\newblock {\em Statistical Science\/}, {\em 17\/}(3), 235--249.

\bibitem[{Brin \& Page(1998)}]{BP98}
Brin, S., \& Page, L. (1998).
\newblock The anatomy of a large-scale hypertextual web search engine.
\newblock {\em Computer Networks and {ISDN} Systems\/}, {\em 30\/}(1-7),
  107--117.

\bibitem[{Brockett(1981)}]{Bro81}
Brockett, P.~L. (1981).
\newblock A note on the numerical assignment of scores to ranked
  categorical-data.
\newblock {\em Journal of Mathematical Sociology\/}, {\em 8\/}(1), 91--101.

\bibitem[{Brockett et~al.(2002)Brockett, Derrig, Golden, Levine, \&
  Alpert}]{BDGLA02}
Brockett, P.~L., Derrig, R.~A., Golden, L.~L., Levine, A., \& Alpert, M.
  (2002).
\newblock Fraud classification using principal component analysis of {RIDIT}s.
\newblock {\em Journal of Risk and Insurance\/}, {\em 69\/}(3), 341--371.

\bibitem[{Brockett \& Levine(1977)}]{BL77}
Brockett, P.~L., \& Levine, A. (1977).
\newblock Characterization of {RIDIT}s.
\newblock {\em Annals of Statistics\/}, {\em 5\/}(6), 1245--1248.

\bibitem[{Bross(1958)}]{Bro58}
Bross, I. (1958).
\newblock How to use {RIDIT} analysis.
\newblock {\em Biometrics\/}, {\em 14\/}(1), 18--38.

\bibitem[{Chakrabarti et~al.(1998)Chakrabarti, Dom, \& Indyk}]{CDI98}
Chakrabarti, S., Dom, B., \& Indyk, P. (1998).
\newblock Enhanced hypertext categorization using hyperlinks.
\newblock In {\em Proceedings of the International Conference on Management of
  Data\/}, (pp. 307--318).

\bibitem[{Demsar et~al.(2004)Demsar, Zupan, Leban, \& Curk}]{DZLC04}
Demsar, J., Zupan, B., Leban, G., \& Curk, T. (2004).
\newblock Orange: From experimental machine learning to interactive data
  mining.
\newblock In {\em Knowledge Discovery in Databases: {PKDD} 2004\/}, (pp.
  537--539). Springer Berlin, Heidelberg.

\bibitem[{Domingos \& Richardson(2001)}]{DR01}
Domingos, P., \& Richardson, M. (2001).
\newblock Mining the network value of customers.
\newblock In {\em Proceedings of the International Conference on Knowledge
  Discovery and Data Mining\/}, (pp. 57--66).

\bibitem[{Eppstein \& Wang(2002)}]{EW02}
Eppstein, D., \& Wang, J. (2002).
\newblock A steady state model for graph power laws.
\newblock In {\em Proceedings of the International Workshop on Web Dynamics\/}.

\bibitem[{Estevez et~al.(2006)Estevez, Held, \& Perez}]{EHP06}
Estevez, P.~A., Held, C.~M., \& Perez, C.~A. (2006).
\newblock Subscription fraud prevention in telecommunications using fuzzy rules
  and neural networks.
\newblock {\em Expert Systems with Applications\/}, {\em 31\/}(2), 337--344.

\bibitem[{Freeman(1977)}]{Fre77}
Freeman, L. (1977).
\newblock A set of measures of centrality based on betweenness.
\newblock {\em Sociometry\/}, {\em 40\/}(1), 35--41.

\bibitem[{Freeman(1979)}]{Fre79}
Freeman, L.~C. (1979).
\newblock Centrality in social networks - conceptual clarification.
\newblock {\em Social Networks\/}, {\em 1\/}(3), 215--239.

\bibitem[{Furlan \& Bajec(2008)}]{FB08}
Furlan, S., \& Bajec, M. (2008).
\newblock Holistic approach to fraud management in health insurance.
\newblock {\em Journal of Information and Organizational Sciences\/}, {\em
  32\/}(2), 99--114.

\bibitem[{Ghosh \& Schwartzbard(1999)}]{GS99}
Ghosh, A.~K., \& Schwartzbard, A. (1999).
\newblock A study in using neural networks for anomaly and misuse detection.
\newblock In {\em Proceedings of the Conference on USENIX Security
  Symposium\/}, (pp. 12--12).

\bibitem[{Girvan \& Newman(2002)}]{GN02}
Girvan, M., \& Newman, M. E.~J. (2002).
\newblock Community structure in social and biological networks.
\newblock {\em Proceedings of the National Academy of Sciences {USA}\/}, {\em
  99\/}(12), 7821--7826.

\bibitem[{Holder \& Cook(2003)}]{HC03}
Holder, L.~B., \& Cook, D.~J. (2003).
\newblock Graph-based relational learning: current and future directions.
\newblock {\em SIGKDD Explorations\/}, {\em 5\/}(1), 90--93.

\bibitem[{Hu et~al.(2007)Hu, Liao, \& Vemuri}]{HLV07}
Hu, W., Liao, Y., \& Vemuri, V.~R. (2007).
\newblock Robust anomaly detection using support vector machines.
\newblock In {\em Proceedings of the International Conference on Machine
  Learning\/}.

\bibitem[{Hummel \& Zucker(1983)}]{HZ83}
Hummel, R.~A., \& Zucker, S.~W. (1983).
\newblock On the foundations of relaxation labeling processes.
\newblock {\em {IEEE} Transactions on Pattern Analysis and Machine
  Intelligence\/}, {\em 5\/}(3), 267--287.

\bibitem[{Jensen(1997)}]{Jen97}
Jensen, D. (1997).
\newblock Prospective assessment of {AI} technologies for fraud detection and
  risk management.
\newblock In {\em Proceedings of the {AAAI} Workshop on {AI} Approaches to
  Fraud Detection and Risk Management\/}, (pp. 34--38).

\bibitem[{Jensen(1999)}]{Jen99}
Jensen, D. (1999).
\newblock Statistical challenges to inductive inference in linked data.
\newblock In {\em Proceedings of the International Workshop on Artificial
  Intelligence and Statistics\/}, (pp. 59--62).

\bibitem[{Kempe et~al.(2003)Kempe, Kleinberg, \& Tardos}]{KKT03}
Kempe, D., Kleinberg, J., \& Tardos, E. (2003).
\newblock Maximizing the spread of influence through a social network.
\newblock In {\em Proceedings of the International Conference on Knowledge
  Discovery and Data Mining\/}, (pp. 137--146).

\bibitem[{Kirkos et~al.(2007)Kirkos, Spathis, \& Manolopoulos}]{KSM07}
Kirkos, E., Spathis, C., \& Manolopoulos, Y. (2007).
\newblock Data mining techniques for the detection of fraudulent financial
  statements.
\newblock {\em Expert Systems with Applications\/}, {\em 32\/}(4), 995--1003.

\bibitem[{Kleinberg(1999)}]{Kle99}
Kleinberg, J.~M. (1999).
\newblock Authoritative sources in a hyperlinked environment.
\newblock {\em Journal of the ACM\/}, {\em 46\/}(5), 604--632.

\bibitem[{Kschischang \& Frey(1998)}]{KF98}
Kschischang, F.~R., \& Frey, B.~J. (1998).
\newblock Iterative decoding of compound codes by probability propagation in
  graphical models.
\newblock {\em {IEEE} Journal on Selected Areas in Communications\/}, {\em
  16\/}(2), 219--230.

\bibitem[{Lu \& Getoor(2003{\natexlab{a}})}]{LG03b}
Lu, Q., \& Getoor, L. (2003{\natexlab{a}}).
\newblock Link-based classification using labeled and unlabeled data.
\newblock In {\em Proceedings of the ICML Workshop on the Continuum from
  Labeled to Unlabeled Data in Machine Learning and Data Mining\/}.

\bibitem[{Lu \& Getoor(2003{\natexlab{b}})}]{LG03a}
Lu, Q., \& Getoor, L. (2003{\natexlab{b}}).
\newblock Link-based text classification.
\newblock In {\em Proceedings of the IJCAI Workshop on Text Mining and Link
  Analysis\/}.

\bibitem[{Maxion \& Tan(2000)}]{MT00}
Maxion, R.~A., \& Tan, K. M.~C. (2000).
\newblock Benchmarking anomaly-based detection systems.
\newblock In {\em Proceedings of the International Conference on Dependable
  Systems and Networks\/}, (pp. 623--630).

\bibitem[{Minka(2001)}]{Min01}
Minka, T. (2001).
\newblock Expectation propagation for approximate bayesian inference.
\newblock In {\em Proceedings of the Conference on Uncertanty in Artificial
  Intelligence\/}, (pp. 362--369).

\bibitem[{Neville \& Jensen(2000)}]{NJ00}
Neville, J., \& Jensen, D. (2000).
\newblock Iterative classification in relational data.
\newblock In {\em Proceedings of the Workshop on Learning Statistical Models
  from Relational Data\/}, (pp. 13--20).

\bibitem[{Newman(2003)}]{New03}
Newman, M. E.~J. (2003).
\newblock The structure and function of complex networks.
\newblock {\em SIAM Review\/}, {\em 45\/}(2), 167--256.

\bibitem[{Newman(2008)}]{New08}
Newman, M. E.~J. (2008).
\newblock Mathematics of networks.
\newblock In {\em The New Palgrave Encyclopedia of Economics\/}. Palgrave
  Macmillan, Basingstoke.

\bibitem[{Noble \& Cook(2003)}]{NC03}
Noble, C.~C., \& Cook, D.~J. (2003).
\newblock Graph-based anomaly detection.
\newblock In {\em Proceedings of the ACM SIGKDD International Conference on
  Knowledge Discovery and Data Mining\/}, (pp. 631--636).

\bibitem[{Oh et~al.(2000)Oh, Myaeng, \& Lee}]{OML00}
Oh, H.~J., Myaeng, S.~H., \& Lee, M.~H. (2000).
\newblock A practical hypertext categorization method using links and
  incrementally available class information.
\newblock In {\em Proceedings of the ACM SIGIR International Conference on
  Research and Development in Information Retrieval\/}, (pp. 264--271).

\bibitem[{Perez et~al.(2005)Perez, Muguerza, Arbelaitz, Gurrutxaga, \&
  Martin}]{PMAGM05}
Perez, J.~M., Muguerza, J., Arbelaitz, O., Gurrutxaga, I., \& Martin, J.~I.
  (2005).
\newblock Consolidated tree classifier learning in a car insurance fraud
  detection domain with class imbalance.
\newblock In {\em Proceedings of the International Conference on Advances in
  Pattern Recognition\/}, (pp. 381--389).

\bibitem[{Phua et~al.(2005)Phua, Lee, Smith, \& Gayler}]{PLSG05}
Phua, C., Lee, V., Smith, K., \& Gayler, R. (2005).
\newblock A comprehensive survey of data mining-based fraud detection research.
\newblock {\em Artificial Intelligence Review\/}.

\bibitem[{Quah \& Sriganesh(2008)}]{QS08}
Quah, J. T.~S., \& Sriganesh, M. (2008).
\newblock Real-time credit card fraud detection using computational
  intelligence.
\newblock {\em Expert Systems with Applications\/}, {\em 35\/}(4), 1721--1732.

\bibitem[{Rupnik et~al.(2007)Rupnik, Kukar, \& Krisper}]{RKK07}
Rupnik, R., Kukar, M., \& Krisper, M. (2007).
\newblock Integrating data mining and decision support through data mining
  based decision support system.
\newblock {\em Journal of Computer Information Systems\/}, {\em 47\/}(3),
  89--104.

\bibitem[{Sanchez et~al.(2009)Sanchez, Vila, Cerda, \& Serrano}]{SVCS09}
Sanchez, D., Vila, M.~A., Cerda, L., \& Serrano, J.~M. (2009).
\newblock Association rules applied to credit card fraud detection.
\newblock {\em Expert Systems with Applications\/}, {\em 36\/}(2), 3630--3640.

\bibitem[{Sen \& Getoor(2007)}]{SG07}
Sen, P., \& Getoor, L. (2007).
\newblock Link-based classification.
\newblock Tech. Rep. CS-TR-4858, University of Maryland.

\bibitem[{Sun et~al.(2005)Sun, Qu, Chakravarti, \& Faloutsos}]{SQCF05}
Sun, J., Qu, H., Chakravarti, D., \& Faloutsos, C. (2005).
\newblock Relevance search and anomaly detection in bipartite graphs.
\newblock {\em ACM SIGKDD Explorations Newsletter\/}, {\em 7\/}(2), 48--55.

\bibitem[{Viaene et~al.(2005)Viaene, Dedene, \& Derrig}]{VDD05}
Viaene, S., Dedene, G., \& Derrig, R.~A. (2005).
\newblock Auto claim fraud detection using bayesian learning neural networks.
\newblock {\em Expert Systems with Applications\/}, {\em 29\/}(3), 653--666.

\bibitem[{Viaene et~al.(2002)Viaene, Derrig, Baesens, \& Dedene}]{VDBD02}
Viaene, S., Derrig, R.~A., Baesens, B., \& Dedene, G. (2002).
\newblock A comparison of state-of-the-art classification techniques for expert
  automobile insurance claim fraud detection.
\newblock {\em Journal of Risk and Insurance\/}, {\em 69\/}(3), 373--421.

\bibitem[{Watts \& Strogatz(1998)}]{WS98}
Watts, D.~J., \& Strogatz, S.~H. (1998).
\newblock Collective dynamics of 'small-world' networks.
\newblock {\em Nature\/}, {\em 393\/}(6684), 440--442.

\bibitem[{Weisberg \& Derrig(1998)}]{WD98}
Weisberg, H.~I., \& Derrig, R.~A. (1998).
\newblock Quantitative methods for detecting fraudulent automobile bodily
  injury claims.
\newblock {\em Risques\/}, {\em 35\/}, 75--101.

\bibitem[{Yang \& Hwang(2006)}]{YH06}
Yang, W.~S., \& Hwang, S.~Y. (2006).
\newblock A process-mining framework for the detection of healthcare fraud and
  abuse.
\newblock {\em Expert Systems with Applications\/}, {\em 31\/}(1), 56--68.

\end{thebibliography}
\end{document}